\documentclass{article}
\usepackage{ijcai24}

\usepackage{times}
\usepackage{soul}
\usepackage{url}
\usepackage[hidelinks]{hyperref}
\usepackage[utf8]{inputenc}
\usepackage[small]{caption}
\usepackage{graphicx}
\usepackage{amsmath}
\usepackage{amsthm}
\usepackage{booktabs}
\usepackage{algorithm}
\usepackage{algorithmic}
\usepackage[switch]{lineno}

\usepackage{multirow}
\usepackage[utf8]{inputenc}
\usepackage{CJKutf8}
\usepackage{hyperref}
\usepackage{xcolor}

\title{For the Misgendered Chinese in Gender Bias Research: Multi-Task Learning with Knowledge Distillation for Pinyin Name-Gender Prediction}

\author{
Xiaocong Du
\and
Haipeng Zhang\footnote{Corresponding author.} \\
\affiliations
ShanghaiTech University\\
\emails
\{duxc2023, zhanghp\}@shanghaitech.edu.cn
}

\begin{document}

\maketitle

\begin{abstract}
    Achieving gender equality is a pivotal factor in realizing the UN's Global Goals for Sustainable Development. Gender bias studies work towards this and rely on name-based gender inference tools to assign individual gender labels when gender information is unavailable. However, these tools often inaccurately predict gender for Chinese Pinyin names, leading to potential bias in such studies. With the growing participation of Chinese in international activities, this situation is becoming more severe. Specifically, current tools focus on pronunciation (Pinyin) information, neglecting the fact that the latent connections between Pinyin and Chinese characters (Hanzi) behind convey critical information. As a first effort, we formulate the Pinyin name-gender guessing problem and design a Multi-Task Learning Network assisted by Knowledge Distillation that enables the Pinyin embeddings in the model to possess semantic features of Chinese characters and to learn gender information from Chinese character names. Our open-sourced method surpasses commercial name-gender guessing tools by 9.70\% to 20.08\% relatively, and also outperforms the state-of-the-art algorithms.
\end{abstract}

\section{Introduction}
Gender equality is an important pillar of the UN's Global Goals for Sustainable Development and numerous gender bias studies work towards this goal. For these studies, not all times is gender information of the target population available, and employing a name-based gender inference tool to assign gender labels based on individual names is a popular solution. Such tools have been widely applied in various contexts, such as understanding the impact of gender composition on teams~\cite{yang2022gender} and revealing gender inequalities among scientists~\cite{lariviere2013bibliometrics,huang2020historical}, academic publications~\cite{dworkin2020extent,squazzoni2021peer}, and in faculty positions~\cite{spoon2023gender}, as shown in Tabel~\ref{tab:publication}.

\begin{figure}[t]
	\centering
	\includegraphics[width=0.4\textwidth]{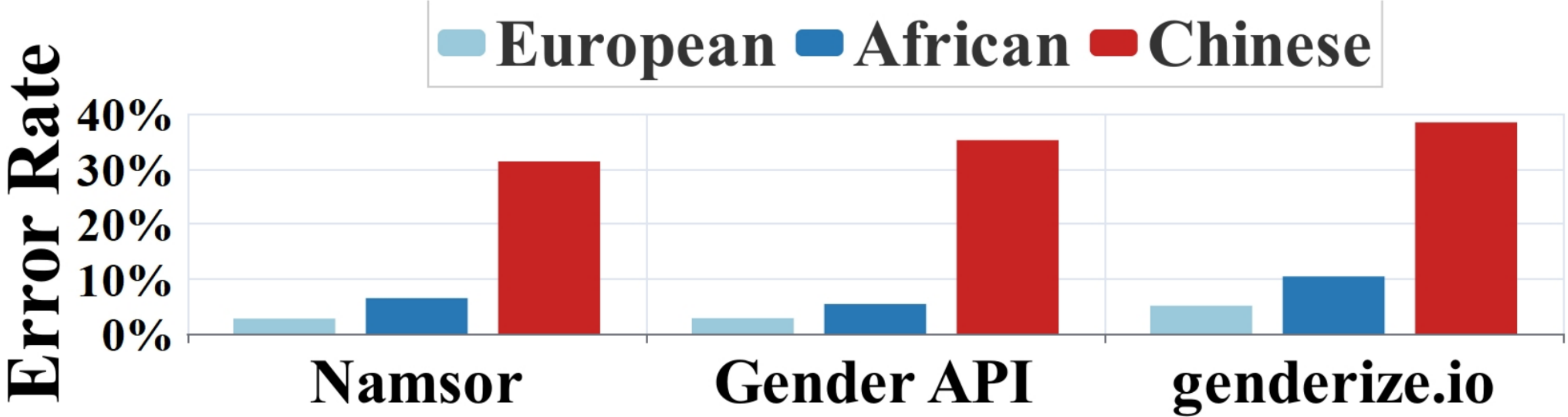}
	\caption{The gender classification error rates of the commercial gender detection tools for names of different origins.}
	\label{fig:intro}
\end{figure}

However, as evidenced in Figure~\ref{fig:intro}, popular name-gender guessing tools, namely Namsor\footnote{\url{https://namsor.app/}}, Gender API\footnote{\url{https://gender-api.com/}}, and genderize.io\footnote{\url{https://genderize.io/}}, perform poorly for Chinese names in Pinyin, which represents the romanized pronunciations of corresponding Chinese characters (Hanzi). The error rate in gender classification for Pinyin names is 7 to 12 times higher than that for European names and for the web service genderize.io, the error rate is as high as 38.39\%. Some studies~\cite{dworkin2020extent,squazzoni2021peer,yang2022gender} use these results for the Chinese in their datasets, while some~\cite{huang2020historical} decide to exclude these samples representing Chinese people from the analysis. Both practices seem problematic: if Chinese Pinyin results are kept, the low accuracy would shake the reliability of the analysis; if they are discarded, there is inevitably a bias since Chinese people are underrepresented. This problem can be more severe, as Chinese people take up considerably large portions of the datasets and there is a growing trend -- for instance, in 2002, Chinese authors write 4.72\% of international papers and this ratio grows to 19.70\% in 2022\footnote{\url{https://www.scimagojr.com/countryrank.php}}. Manual verification by searching for scientists' homepages may produce accurate results~\cite{lariviere2013bibliometrics}, while it is not practical for datasets containing millions of individuals without homepages.

\begin{table*}[t]
	\centering
	\begin{tabular}{ccccc}
		\toprule
	Study & Venue & Data Source & Gender Inferring Tool & Pinyin names excluded?\\
		\midrule
	\citeauthor{huang2020historical}~\shortcite{huang2020historical}  & PNAS & Web of Science & genderize.io & Yes \\
    \citeauthor{dworkin2020extent}~\shortcite{dworkin2020extent} & Nature Neuroscience & Web of Science & genderize.io & No \\
	\citeauthor{squazzoni2021peer}~\shortcite{squazzoni2021peer} & Science Advances & 145 journals & Gender API & No
 \\
    \citeauthor{yang2022gender}~\shortcite{yang2022gender}& PNAS & Microsoft Academic Graph & Namsor  & No \\
    \citeauthor{lariviere2013bibliometrics}~\shortcite{lariviere2013bibliometrics} & Nature & Web of Science &  manually annotated & No \\
		\bottomrule
	\end{tabular}
	\caption{Instances of gender bias research that infer genders from romanized names.}
	\label{tab:publication}
\end{table*}

\begin{figure}[t]
	\centering
	\includegraphics[width=0.45\textwidth]{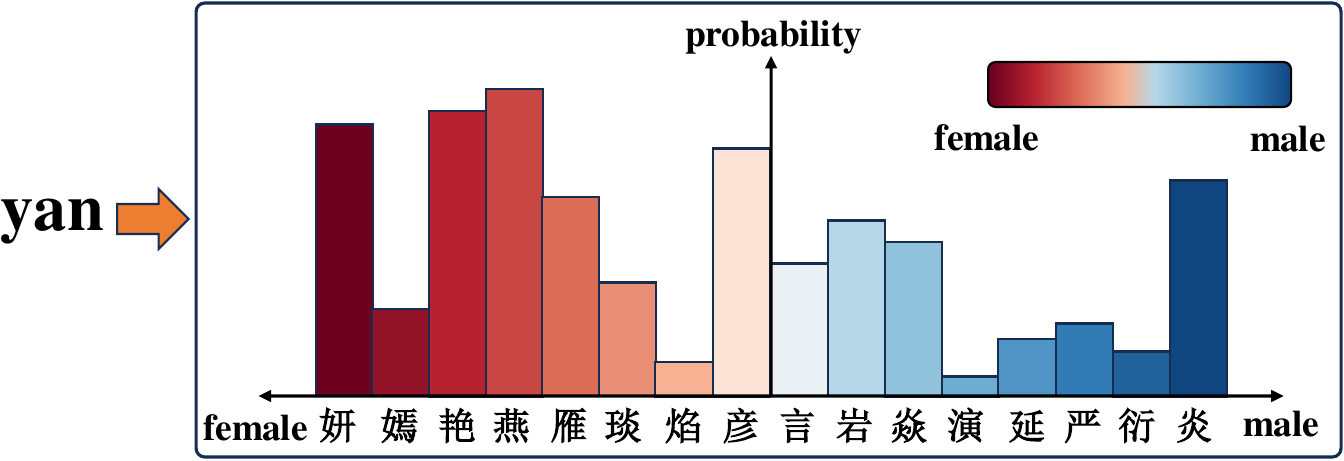}
	\caption{The possible correspondences between Pinyin and Chinese characters, as well as the gender information embedded in Chinese characters names, take `yan' as an example. All the Chinese characters above have the same Pinyin `yan'.}
	\label{fig:pinyin}
\end{figure}

In most Western languages, gender information is encapsulated within the names that consist of letters. In languages such as English or German, there exist distinct lists of male and female names, and the gender tendency of a name can be easily computed as the ratios of males/females under that name. In languages such as Italian, gender can be marked on a specific part of a name (e.g., suffixes like `a' for females or `o' for males). The former is categorized as the conventional type of gender marking languages, while the latter is classified as the formal type~\cite{ackermann2021sound}.

Pinyin only represents the pronunciations of the Chinese characters or in other words, Chinese characters are converted into Pinyin. Pronunciations, convey some helpful information for gender guessing~\cite{10.1609/aaai.v37i12.26688,jia2019gender} but we have to note that the Chinese characters behind, are more informative~\cite{10.1609/aaai.v37i12.26688}. If it were one-to-one mapping from Pinyin to Chinese characters, it would be a (almost) solved problem since we can infer genders from names in Chinese characters with state-of-the-art tools such as CHGAT~\cite{10.1609/aaai.v37i12.26688}. Unfortunately, Pinyin can be associated with different Chinese characters based on varying probabilities, as shown in Figure~\ref{fig:pinyin}.

Chinese represents the third type separated from the two aforementioned language types, namely, the semantic type. In Chinese names, gender is indicated by the semantics of the name.\begin{CJK}{UTF8}{gbsn} For example, at the left end of the horizontal axis in Figure~\ref{fig:pinyin}, the character `妍' connotes `beautiful', aligning with the female gender stereotype, hence it is commonly used in female names. Conversely, at the other end, `炎' signifies `flame of fire', which relates to the male gender stereotype, making it prevalent in male names. Characters `妍' and `炎' exhibit distinct gender tendencies.\end{CJK} However, when they are converted to their Pinyin forms `yan', they become indistinguishable. Pinyin only represents the pronunciation of the characters, resulting in the loss of the tonal, symbolic, and semantic information. This transition leads to the loss of gender information embedded in the Chinese names. Given that characters are pivotal in inferring gender in Chinese names, previous methods~\cite{hu2021s,van2023open} overlook the potential link between Pinyin and Chinese characters.

We therefore want to model the uncertainty in the connections between Pinyin and Chinese characters and utilize the gender information in Pinyin and more importantly, Chinese characters. A straightforward approach to establish the Pinyin-character connection is converting Pinyin into the most possible Chinese character. However, this results in the loss of information from other possible characters. Instead, we introduce a Pinyin-to-character prediction task as an intermediate auxiliary task to learn the associations between Chinese characters and Pinyin, as well as the relationships among different characters, particularly those sharing the same Pinyin. Since our main task is Pinyin name-gender prediction task, we employ Multi-Task Learning (MTL) to integrate two tasks into a cohesive framework. By allowing two tasks to share encoder parameters, the model can learn external knowledge from the supervision signals of characters and generate characters enhanced embeddings. Consequently, the Pinyin embeddings learned from multi-task learning essentially integrate the information of all Chinese characters associated with that Pinyin, while preserving the one-to-many relationship between Pinyin and Chinese characters.

The MTL module mainly captures the Pinyin gender tendency and the association between Pinyin and Chinese characters, represented by the vertical dimension in Figure~\ref{fig:pinyin}. However, the gender-related information embedded within the Chinese characters, i.e., the horizontal dimension in Figure~\ref{fig:pinyin}, remains underexplored. Though we can assign binary gender labels to the embeddings learned from the MTL step, this only provides gender information at sample level and does not encompass the overall gender tendency of the Chinese characters behind each Pinyin. The gender tendency of each character can be learned by training a Chinese character name-gender prediction model. We then want to seamlessly transfer this learned information to the embeddings that model the Pinyin-character connections from MTL. This can be naturally done by Knowledge Distillation that encapsulates the knowledge learned from a teacher model (i.e., Chinese character name-gender prediction model here) in the form of soft targets, which are used to supervise the training of a student model (i.e., Pinyin name-gender prediction model here). By performing feature and response based Knowledge Distillation between two models, the gender tendency in Pinyin embeddings becomes clearer under the guidance of gender-specific characteristics of Chinese character names. 

We summarize our contributions as follows:
\begin{enumerate}
	\item We bring the problem of misgendering or excluding Chinese individuals in current gender bias studies to attention, which is attributed to the poor performance of commercial name-gender inference tools designed for Western names. Accordingly, we define the new and important task of Chinese Pinyin name-gender prediction.
	\item We develop a knowledge distillation assisted multi-task learning network for this task to model the connection between the Pinyin and Chinese character forms of Chinese names. Our method outperforms commercial gender inferring services by 9.70\% to 20.08\% relatively and it is open-sourced\footnote{\url{https://github.com/PinyinNameGenderPrediction/pinyin_name_gender_prediction}}.
	\item We determine the potential upper bound of accuracy in the task of Pinyin name-gender guessing. For names in other languages (e.g., Japanese) that are romanized in a `many-to-one' fashion, our model is a possible solution.
\end{enumerate}

\section{Related Work}

\subsection{Name-based Gender Prediction}
Many efforts have been dedicated to enhancing name-based gender prediction tools for better service in scientific research. \citeauthor{santamaria2018comparison}~\shortcite{santamaria2018comparison} evaluate the performance of five frequently used paid web services for gender inference. \citeauthor{hu2021s}~\shortcite{hu2021s} develop character-level machine learning and deep learning methods for gender prediction tasks based on English names. \citeauthor{van2023open}~\shortcite{van2023open} propose a Cultural Consensus Theory (CCT) model that can unify data from multiple sources and offer a taxonomy of names to assess dataset composition. According to their analysis, the performance of current name-gender methods is nearing its ceiling. They believe that potential improvements can only come from increasing data volume and enhancing methodological transparency. However, this conclusion is somewhat one-sided, as the authors focus primarily on languages like English and other Latin-based languages, overlooking the nuances of Chinese names. 

Chinese names differ from English names in that they possess richer lexical meanings, often gender-related. Additionally, Chinese characters provide more gender clues compared to English letters. In Chinese, leveraging the inherent information in names to improve gender prediction performance is feasible. Ngender\footnote{\url{https://github.com/observerss/ngender}} determines the gender probability of a first name by calculating the product of the gender probabilities of each letter. PBERT~\cite{jia2019gender} learns gender characteristics of Chinese names through character embeddings and pronunciation embeddings of Chinese names. CHGAT~\cite{10.1609/aaai.v37i12.26688} represents Chinese character names by constructing a heterogeneous graph attention network of Chinese character radicals and pronunciations. While progress has been made in gender prediction for Chinese names in character form, in more international contexts, Chinese names are often presented in Pinyin form. To our knowledge, there has yet to be research on the task of gender classification for Pinyin names. Multilingual language models may handle different languages and their transliterated versions. However, this transfer relies on language structure~\cite{pires2019multilingual}, which does not exist in extremely short texts such as names.

\subsection{Deep Multi-task Learning}
The basic idea of multi-task learning is that knowledge acquired from one task may prove beneficial when performing another related task~\cite{crawshaw2020multi}. Traditional multi-task learning architectures are quite simplistic:  All tasks share a global feature encoder through either hard or soft parameter sharing, with each task having its separate output branch. Decoder-based architecture~\cite{xu2018pad} introduce intermediate auxiliary prediction tasks in a multi-task network to extract features that have been disentangled according to the output tasks, making it easier for other tasks to extract relevant information \cite{vandenhende2020revisiting}. Recently, \citeauthor{yang2022effectiveness}~\shortcite{yang2022effectiveness} employ a multi-task network in automatic speech recognition to jointly learn the features of Pinyin and Chinese characters, demonstrating that Pinyin and Chinese characters can mutually enhance each other. As they treat Pinyin and Chinese characters equally, in our scenario, we only concern with the performance of Pinyin-related task. We initially facilitate knowledge transfer between Pinyin gender prediction task and Pinyin-to-Character prediction task through hard parameter sharing. Here Pinyin-to-Character prediction task serves as an auxiliary task to enhance our main task. It is known that a suitable auxiliary task for the target task can reap the benefits of multi-task learning~\cite{ruder2017overview}. We then distill the gender-specific features from ground truth Chinese character names instead of from intermediate predictions as in~\cite{xu2018pad}.

\subsection{Knowledge Distillation}
Knowledge distillation transfers knowledge from a large teacher model to a smaller student model, enabling the smaller model to achieve performance close to that of the larger model. \citeauthor{hinton2015distilling}~\shortcite{hinton2015distilling} first propose response-based knowledge distillation. It views the output of the teacher network's logits layer as soft labels containing additional information, which are used to supervise the learning of the student model. Feature-based knowledge distillation works on the principle of feature alignment. It is first introduced in FitNet~\cite{romero2014fitnets}. Though knowledge distillation is traditionally used for model compression, it can also be viewed as a way of transferring knowledge across domains~\cite{gou2021knowledge}. In our scenario, we set the Chinese character names gender prediction network as the teacher model and the Pinyin name-gender prediction network as the student model. The transfer of knowledge from Chinese characters to Pinyin is achieved through knowledge distillation between the two models.

\begin{figure*}[ht]
	\centering
    \hspace*{0.2cm}
	\includegraphics[width=0.95\textwidth]{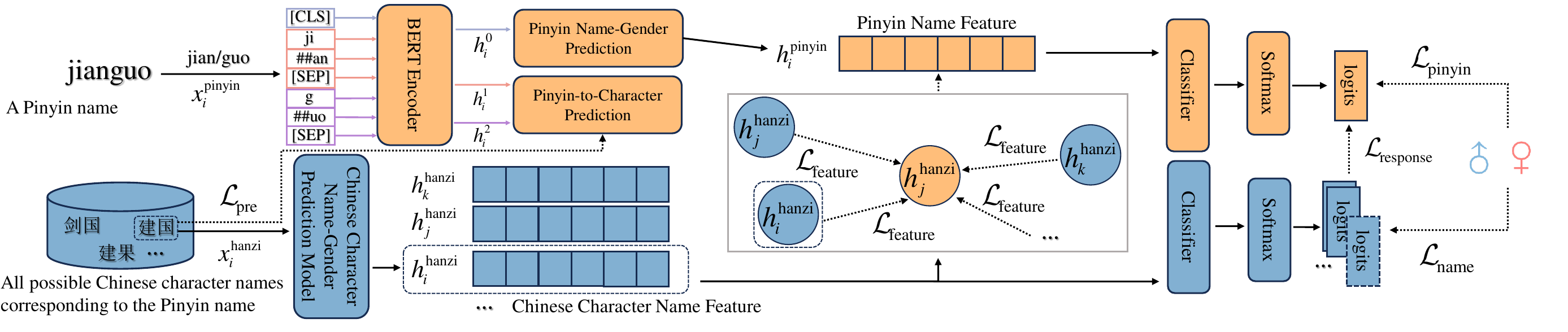}
	\caption{System structure. Orange represents the Pinyin name-gender prediction model (i.e., the student model) that incorporates the Multi-Task Learning module, while blue denotes the Chinese character name-gender prediction model (i.e., the teacher model).}
	\label{fig1}
\end{figure*}

\section{Problem Statement}

We denote the Pinyin form of Chinese names as $X_{\textrm{pinyin}}$, whereas the logographic (Chinese character) representation is indicated as $X_{\textrm{hanzi}}$. Given a training dataset $D=\{X_{\textrm{pinyin}},X_{\textrm{hanzi}},y\}$, our main goal is to train a model $f_1$ with trainable parameters $\theta_1$ to classify $X_{\textrm{pinyin}}$ into binary category $y\in\{0,1\}$:

\begin{equation}
	f_1(X_{\textrm{pinyin}},\theta_1) \rightarrow y
\end{equation}

Without loss of generality, we let $y=1$ denote female and $y=0$ signify male. To capitalize on supplementary information contained within the Chinese names in character form, we develop two additional models $f_2$ and $f_3$ to serve as auxiliaries to the main model $f_1$:

\begin{equation}
	f_2(X_{\textrm{hanzi}},\theta_2) \rightarrow y
\end{equation}
\begin{equation}
	f_3(X_{\textrm{pinyin}},\theta_3) \rightarrow X_{\textrm{hanzi}}
\end{equation}

Specifically, $f_2$ is a Chinese character name-gender prediction model, whereas $f_3$ focuses on predicting Chinese character names based on their Pinyin form.

\section{Method}
The overall structure of our model is shown in Figure~\ref{fig1}. First, we preprocess Pinyin names and feed them into the BERT encoder to generate vector representations. Second, a Pinyin-to-Character prediction task is co-trained with the Pinyin name-gender prediction task, constituting the Multi-Task Learning module (i.e., the orange part) that captures the Pinyin gender tendency and the association between Pinyin and Chinese characters. Third, we design a Knoweledge Distillation module and apply a Chinese character name-gender prediction model (i.e., the blue part) as the teacher model. The output of the last layer and intermediate layers of the teacher model are used to supervise the training of the Pinyin name-gender prediction model, capturing the gender-related information embedded within the Chinese characters. Finally, the distilled information is used for gender classification. The input of our model is Pinyin names and Chinese character names during training. While during the testing phase, the model solely utilizes Pinyin names for gender prediction. We detail the design of our model in the following sections.

\subsection{Pinyin Preprocessing}
Since we want to use BERT as our encoder, the default tokenizer will erroneously separate letters that do not belong to a Pinyin syllable. \begin{CJK}{UTF8}{gbsn} Taking the two-character name `建国' (jianguo) as an example, the default tokenizer separates `jianguo' into `ji', `ang' and `uo', while `建国' actually corresponds to the Pinyin syllables `jian' and `guo'.\end{CJK} To avoid this, we use pinyinsplit\footnote{\url{https://github.com/llb0536/PinyinSplit}} to segment Pinyin names into syllables that correspond to individual Chinese characters.

\subsection{Pinyin-to-Character Prediction}
To obtain a Pinyin representation enriched with information from Chinese characters, we introduce a Pinyin-to-Character Prediction task as an auxiliary task for Multi-Task Learning. For an input Pinyin name $x^{\textrm{pinyin}}_i \in X_{\textrm{pinyin}}$, taking a two-character name as an example, we first segment the letter string $x^{\textrm{pinyin}}_i$ into Pinyin syllables corresponding to Chinese characters $P_i=\{p_i^1,p_i^2\}$. Then, we feed the split Pinyin name into BERT encoder and get a Pinyin representation $H_i^{\textrm{pinyin}} = \{h_i^0,h_i^1,h_i^2\}$. Here, $h_i^0$ represents the feature of the entire Pinyin name, which is used for subsequent gender classification. While $h_i^1$ and $h_i^2$ represent the features of individual syllables, used for Chinese character name prediction. The target of the Pinyin-to-Character prediction task is $x^{\textrm{hanzi}}_i = \{c_i^1,c_i^2\} \in X_{\textrm{hanzi}}$. The prediction loss of this task is:

\begin{equation}
	\mathcal{L}_{\textrm{pre}}=\frac{1}{N}\sum\limits_{i}{-[\log P\left( c_{i}^{1}|\{h_{i}^{1},h_{i}^{2}\} \right)+\log P(c_{i}^{2}|\{h_{i}^{1},h_{i}^{2}\})]}
\end{equation}

\subsection{Response-Based and Feature-Based Knowledge Distillation}
The Pinyin representation learns the connection between Pinyin and Chinese characters from the Pinyin-to-Character prediction task, but the gender-related features in Chinese characters should be given more attention. To acquire the gender features of Chinese character names, we establish a Chinese character name-gender prediction task. Another BERT encoder and a single layer fully connected network extract gender information from the Chinese characters. For an input Chinese character name $x^{\textrm{hanzi}}_i = \{c_i^1,c_i^2\} \in X_{\textrm{hanzi}}$ and the gender label $y_i$ , the prediction loss of this task is:

\begin{equation}
	\mathcal{L}_{\textrm{name}}=\frac{1}{N}\sum\limits_{i}{-\log P\left(y_i|\{c_{i}^{1},c_{i}^{2}\} \right)}
\end{equation}

To transfer the gender information of Chinese character names to Pinyin names, we perform the feature-based and response-based distillation between the Chinese character model and the Pinyin model. Specifically, we use the features and the logits of Chinese character names to supervise the training of the Pinyin names. In Chinese character model, we denote the output of the BERT encoder as $h_i^{\textrm{hanzi}}$ and the output of the last fully connected layer as logits vector $z_i^{\textrm{hanzi}}$. In Pinyin model, a linear layer projects Pinyin name representation $h_i^{0}$ to $h_i^{\textrm{pinyin}}$, while another generates the logits of Pinyin name $z_i^{\textrm{pinyin}}$ from  $h_i^{\textrm{pinyin}}$. The distillation loss for feature-based knowledge transfer is defined as:

\begin{equation}
	\mathcal{L}_{\textrm{feature}}=\frac{1}{N}\sum\limits_{i}{\|h_i^{\textrm{hanzi}},h_i^{\textrm{pinyin}}\|_2}
\end{equation}

The distillation loss for response-based knowledge transfer is defined as:

\begin{align}
	p({{z}_{i}}) &= \frac{\exp ({{z}_{i}})}{\sum\nolimits_{j}{\exp ({{z}_{j}})}} \\
    \mathcal{L}_{\textrm{response}} &= \frac{1}{N}\sum\limits_{i}{D_{\textrm{KL}}(p(z_i^{\textrm{pinyin}}) || p(z_i^{\textrm{hanzi}}))}
\end{align}

Then, the loss function of Pinyin name-gender prediction with response-level and feature-level knowledge distillation can be defined as:

\begin{equation}
	\mathcal{L}_{\textrm{pinyin}}=\frac{1}{N}\sum\limits_{i}{-\log P\left(y_i|z_i^{\textrm{pinyin}} \right)}
\end{equation}

Finally, we combine all the loss functions to formulate our ultimate objective function:

\begin{equation}
	\mathcal{L} = \mathcal{L}_{\textrm{pre}} + \mathcal{L}_{\textrm{name}} + \mathcal{L}_{\textrm{feature}} + \mathcal{L}_{\textrm{response}} + \mathcal{L}_{\textrm{pinyin}}
\end{equation}

\section{Experiments}

\subsection{Experimental Setup}

\subsubsection{Datasets}
We evaluate our model with three distinct datasets. The first is a dataset of 9,800 names compiled from the 2005 China’s 1\% Population Census~\cite{bao2021novel}. The second encompasses a collection of 20,000 names from several public sources in China~\cite{sebo2022accurate}. To facilitate a fair comparison with paid web services that rely on large-scale data, we utilized a third dataset comprising 58,393,173 records collected from Guangdong province government~\cite{10.1609/aaai.v37i12.26688}, named as 58M Names dataset. The detail information of the datasets is shown in Table~\ref{table1}.

\begin{table}[h]
	\centering
	\begin{tabular}{ccccc}
		\toprule
		& Records & UH Name & UP Name\\
		\midrule
		9,800 Names & 9,800 & 6,972 & 4,203\\
		20,000 Names & 20,000 & N/A & 17,948\\
		58M Names & 58,393,173 & 560,706 & 65,971\\
		\bottomrule
	\end{tabular}
	\caption{Statistics of the datasets, where UH Name means Unique Hanzi (Chinese Character) first name and UP Name means Unique Pinyin first name. For the 20,000 Names dataset, it lacks the first name in Chinese character form.}
	\label{table1}
\end{table}

\subsubsection{Train/Test Split}
The 9,800 Names dataset is split into 8:1:1 for training, validation, and test. For a fair comparison with paid web services, we train our model using a significantly larger 58M Names dataset. Subsequently, we conduct testing across the entire scope of the 9,800 Names dataset and 20,000 Names dataset.

\subsubsection{Evaluation Metrics}
The traditional metrics are Accuracy (Acc), Precision (P), Recall (R), and F1-score (F1). While in our scenario, we are not assessing the outcomes of a strict binary classification problem. Besides `male' and `female', an `unknown' category can also be returned by paid web services. To comprehensively evaluate the performance of different methods, we adopt a evaluation metric \cite{wais2016gender} that incorporates the `unknown' category. This metric has already been utilized in evaluating the performance of paid web services \cite{santamaria2018comparison,sebo2022accurate}. The new confusion matrix is defined in Table~\ref{table2}.

Elements $m_m$ and $f_f$ are the correct classifications, while elements $m_f$ and $f_m$ are misclassifications. The sum of both can be simply referred to as classifications, and elements $m_u$ and $f_u$ represent non-classifications. Both misclassifications and non-classifications are included under the term inaccuracies. Based on the confusion matrix, four performance metrics are introduced:

\begin{subequations}
	\begin{align}
		& \textrm{errorCoded} = \frac{f_m+m_f+m_u+f_u}{m_m+f_f+m_f+f_m+m_u+f_u} \\
		& \textrm{errorCodedWithoutNA} = \frac{f_m+m_f}{m_m+f_f+m_f+f_m} \\
		& \textrm{naCoded} = \frac{m_u+f_u}{m_m+f_f+m_f+f_m+m_u+f_u} \\
		& \textrm{errorGenderBias} = \frac{m_f-f_m}{m_m+f_f+m_f+f_m}
	\end{align}
\end{subequations}

The first metric, errorCoded, treats both misclassifications and non-classifications as errors, thereby measuring the overall performance of the method. The second metric, errorCodedWithoutNA, assesses the errors excluding the `unknown' category, proving useful when high predictive accuracy is demanded and a significant number of omissions is acceptable. The third metric, naCoded, quantifies the rate of missed outcomes. The last metric, errorGenderBias, estimates the direction of bias in gender prediction and is used to assess whether errors are more frequent with male or female names. A positive value indicates that males are misclassified more frequently than females, and vice versa.

\begin{table}[t]
	\centering
	\begin{tabular}{|c|c|c|c|c|}
		\hline
		\multicolumn{2}{|c|}{\multirow{2}{*}{confusion matrix}} & \multicolumn{3}{|c|}{predicted class} \\
		\cline{3-5}
		\multicolumn{2}{|c|}{} & male & female & unknown \\
		\hline
		\multirow{2}{*}{true class} & male & $m_m$ & $m_f$ & $m_u$\\
		\cline{2-5}
		& female & $f_m$ & $f_f$ & $f_u$\\
		\hline
	\end{tabular}
	\caption{Confusion matrix of six possible classification outcomes.}
	\label{table2}
\end{table}

\begin{table}[t]
	\centering
	\begin{tabular}{ccccc}
		\toprule
		Methods & P & R & F1 & Acc\\
		\midrule
		CCT (NA=`male') & 0.6431 & 0.6133 & 0.5880 & 0.6071 \\
		CCT (NA=`female') & 0.6764 & 0.6604 & 0.6550 & 0.6643 \\
        GPT3.5 & 0.6948 & 0.6191 & 0.5781 & 0.6191 \\
        Conversion & 0.6824 & 0.6825 & 0.6824 & 0.6827\\
		char-BERT & 0.7142 & 0.7126 & 0.7117 & 0.7122 \\
		Pinyin-Ngender & 0.7158 & 0.7157 & 0.7154 & 0.7155 \\
		PBERT (Pinyin only) & 0.7180 & 0.7175 & 0.7169 & 0.7171  \\
		Ours & \textbf{0.7349} & \textbf{0.7341} & \textbf{0.7334} & \textbf{0.7337} \\
		\bottomrule
	\end{tabular}
	\caption{Performance comparison for models trained and tested on the 9,800 Names dataset with 5-fold cross validation. GPT3.5 is given the prompt and the 9,800 names.}
	\label{table3}
\end{table}

\begin{table*}[t]
	\centering
	\begin{tabular}{cccccc}
		\toprule
		& Methods & errorCoded & errorCodedWithoutNA & naCoded & errorGenderBias \\
		\midrule
		\multirow{4}{*}{9,800 Names} & Gender API & 0.3540 & 0.3030 & 0.0732 & -0.1588 \\
		& genderize.io & 0.3425 & 0.3156 & 0.0393 & -0.1130 \\
		& Namsor & 0.3133 & 0.3133 & 0.0000 & -0.0671 \\
		& Ours & \textbf{0.2829} & \textbf{0.2829} & 0.0000 & -0.1196 \\
		\midrule
		\multirow{4}{*}{20,000 Names} & Gender API & 0.6496 & 0.4183 & 0.3976 & -0.0906 \\
		& genderize.io & 0.6124 & 0.4306 & 0.3193 & -0.0438 \\
		& Namsor & 0.5253 & 0.4065 & 0.2001 & -0.2021 \\
		& Ours & \textbf{0.3847} & \textbf{0.3847} & 0.0000 & -0.0862 \\
		\bottomrule
	\end{tabular}
	\caption{Performance comparison with paid web services. Models are trained on the 58M dataset.}
	\label{table4}
\end{table*}

\subsubsection{Baselines}
We compare our method with nine baselines. In the first baseline, we directly convert Pinyin names to Chinese characters. The second one, char-BERT, is a character-based name-gender prediction model tailored for alphabetic writing systems. The subsequent two (Ngender and PBERT) are representative methods for predicting gender from Chinese character names. We modify these models to work on Pinyin names. We also compare our method with three popular name-based gender inferring tools provided by paid web services, as well as the current most popular large language model, GPT-3.5\footnote{\url{https://platform.openai.com/docs/models/gpt-3-5-turbo}}. The last CCT is a recently proposed dictionary-based method for name-based gender classification.
\begin{itemize}
    \item \textbf{Conversion:} It simply converts Pinyin names into most possible Chinese character names and employ a SOTA  Chinese character names gender classification tool CHGAT~\cite{10.1609/aaai.v37i12.26688} to predict genders from these Chinese character names.
	\item \textbf{char-BERT~\cite{hu2021s}:} This model segments names into individual letters (i.e., characters) and generates a vector representation for each letter. It captures letter-level information within names and performs well with English names.
	\item \textbf{Pinyin-Ngender:} Ngender is a commonly used Chinese character name-gender prediction tool based on the Naive Bayes. We implement a Pinyin version of Ngender, called Pinyin-Ngender, by replacing the Chinese characters with their corresponding Pinyin.
	\item \textbf{PBERT~\cite{jia2019gender}:} Pinyin BERT (PBERT) employs a pre-trained BERT to generate vector representations for names in both their Chinese character and Pinyin forms, using their concatenation for gender classification. Here, we implement its Pinyin-only version.
	\item \textbf{Paid Web Services:} Namsor, Gender API and genderize.io are mainstream paid web services which perform well on European names with wide applications in scientific research~\cite{santamaria2018comparison,van2023open}.
    \item \textbf{GPT-3.5:} GPT-3.5 has demonstrated remarkable capabilities in name-gender tasks~\cite{michelle2023gender}. Here we use this prompt: ``I will give you the first names of some people, please infer their gender, use 0 to represent male and 1 to represent female.''.
	\item \textbf{CCT~\cite{van2023open}:} The CCT model is based on Cultural Consensus Theory. It forms a gender consensus for names from different data sources by iteratively calculating the capability of each data source and the collective estimation of all sources on the name. Since the original CCT model is based on datasets from 36 sources collected by its authors, and our data sources are far fewer, we only implement a simplified version of this model. Due to CCT's inability to assign genders to names not present in the training data, we are compelled to uniformly designate all instances of missing gender information (NA) as either `male' or `female', denoted as CCT (NA=`male') and CCT (NA=`female').
\end{itemize}

\begin{figure}[t]
	\centering
	\includegraphics[width=0.45\textwidth]{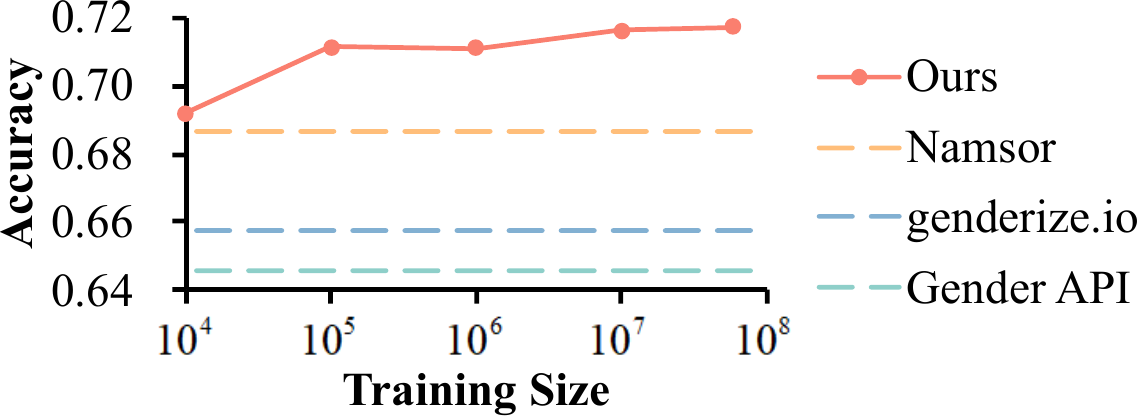}
	\caption{Model performance under different training set sizes. (The x-axis is in log scale.)}
	\label{fig:dis}
\end{figure}

\subsection{Experimental Results}
Table~\ref{table3} shows the average results of 5-fold cross-validation on the 9,800 Names dataset beyond paid services since they cannot be evaluated using traditional metrics. Among all these baselines, our method achieves the best overall performance and exhibits a relative improvement of 2.31\% in accuracy compared to the best baseline model, which suggests the effectiveness of incorporating Chinese character information into the Pinyin name-gender prediction task. The observation that directly converting Pinyin into Chinese characters performs even worse than methods only using Pinyin underscores the complexity of the one-to-many relationship between Pinyin and Chinese characters. We also notice that the syllable-baesd method, PBERT (Pinyin only) and Pinyin-Ngender, perform better than character-based method, char-BERT, which indicates that Pinyin syllables convey gender information. GPT-3.5 does not perform very well on this task, compared with specifically designed models.

Paid web services typically rely on large-scale datasets. Given that we start with a dataset of 9,800 names, which is significantly smaller than the datasets used by paid web services, we want to see what happens as the size of the training set continues to increase. In each iteration, we randomly select a subset of names from the 58M Names dataset for training and evaluate the performance of our model on the entire 9,800 Names dataset. We report the experimental results in Figure~\ref{fig:dis}. When the training size is 10,000, which is approximately equal to that of the test set, our method has surpassed the best commercial gender detection tools. As the size of the training set continues to increase, our method performs increasingly better compared to paid web services.

Finally, we train our model utilizing the entire 58M Names dataset and evaluate on two distinct datasets representing scenarios with and without Chinese character first names. The experimental results are presented in Table~\ref{table4}. On the 9,800 Names dataset, our method outperforms the leading mainstream name-based gender inferring tools, achieving relative improvements of 9.70\% to 20.08\% (i.e., 3.04\% to 7.11\% absolutely) in errorCoded, the proportion of misclassifications and nonclassifications. On the 20,000 Names dataset, the improvements brought by our method were even more significant, achieving relative gains ranging from 26.77\% to 40.78\% (i.e., 14.06\% to 26.49\% absolutely) in errorCoded.

Additionally, we observe that in almost all results of various methods, errorGenderBias is consistently negative. This means that females are more prone to being misclassified compared to males, which could likely be caused by an imbalance in male/female ratio within the training set.

\begin{table}[t]
	\centering
	\begin{tabular}{ccccc}
		\toprule
		Methods & P & R & F1 & Acc \\
		\midrule
		w/o distill\&namepre & 0.7141 & 0.7142 & 0.7142 & 0.7143 \\
		w/o logits\&feat & 0.7253 & 0.7253 & 0.7253 & 0.7255 \\
		w/o logits & 0.7287 & 0.7288 & 0.7285 & 0.7286 \\
		Ours & \textbf{0.7349} & \textbf{0.7341} & \textbf{0.7334} & \textbf{0.7337} \\
		\bottomrule
	\end{tabular}
	\caption{Performance of different variants under the 9,800 dataset.}
	\label{table5}
\end{table}

\subsection{Ablation Study}
To validate the effectiveness of our proposed multi-task learning module and knowledge distillation module, we incrementally remove the response-based knowledge distillation (w/o logits), feature-based knowledge distillation (w/o logits\&feat), the entire teacher network and the task of predicting Chinese character names (w/o distill\&namepre) from our model. We test various variants of our model on the 9,800 Names dataset, and the experimental results are presented in Table~\ref{table5}. Unsurprisingly, our method outperformed all other variants. Furthermore, as response-based knowledge, feature-based knowledge, and the multi-task learning module are removed, the model's accuracy progressively decline. 

\section{Discussion}
\paragraph{Performance Upper Bound.} It is noticed that, as training set size grows, the performance increase of our method slows down. This is consistent with observations in other name-gender studies~\cite{van2023open,lockhart2023name}. It appears that, for name-gender guessing tasks, there exists an information-theoretical limit to accuracy. Beyond this threshold, additional reference data or more advanced modeling techniques may not enhance the performance.

As mentioned in Introduction, if we know the exact Chinese characters behind each Pinyin name, the problem would be almost solved. In the previous work for Chinese character names gender prediction, similarly trained on the 58M dataset and tested on the 9,800 dataset, the accuracy of current SOTA model~\cite{10.1609/aaai.v37i12.26688} reaches 81.47\%. This may be the most ideal scenario for Pinyin name-gender classification, which is (almost) impossible to achieve since it is impossible to determine the exact Pinyin-to-characters mappings. Our proposed method achieves 71.71\% in accuracy (i.e., 28.29\% in errorCoded), which means the potential upper bound for this task may lie between these two extremes and further enhancements towards the upper bound would be increasingly difficult.

\paragraph{Future Directions.} Though this work focuses on gender detection of romanized Chinese names, a similar approach could be applied to other languages that are romanized in English records, such as Japanese in which many-to-one mappings exist. If the conversion of names from different languages into alphabetic script is inevitable, our method can to some extent mitigate the loss of gender information resulting from this transliteration process.

Naming preferences in China change over time and across regions, according to the Ministry of Public Security of China\footnote{\url{https://www.gov.cn/xinwen/2021-02/08/content_5585906.htm}}. It is unclear how these specifics of name-gender dictionaries affect the model performance on test sets from different time and regions and we leave it to future explorations.

\section{Conclusion}

We identify the problem of misgendering Chinese Pinyin names and the exclusion of Chinese individuals from analyses in previous gender bias studies, which is resulted from the low accuracy of name-gender guessing tools on Pinyin names. We define a new Pinyin name-gender guessing task accordingly and design a Multi-Task Learning Network assisted by Knowledge Distillation to transfer semantic and gender information from the rich Chinese characters to Pinyin. This open-sourced method outperforms commercial gender classification tools and other baselines. Aligned with the UN's SDGs to promote gender equality, we hope that it can help consolidate gender bias studies such that Chinese individuals' genders are inferred with higher accuracy and they would not be excluded from these studies. 

\appendix
\section*{Ethical Statement}
Gender bias visibly or invisibly exists in many aspects of our lives and severely hinders humankind’s sustainable future. We design our model to help researchers and policymakers to confirm its systematic existence, to estimate its negative impact, to figure out its mechanism, to educate the public and to develop corresponding scientific and effective policies. However, not all names are strongly gendered. Although a name can serve as a gender characteristic, it does not necessarily reflect the actual sexuality of a person. Model predictions should still be treated with caution.

\bibliographystyle{named}
\bibliography{ijcai24}

\end{document}